\documentclass[conference]{ieeeconf}
\IEEEoverridecommandlockouts
\usepackage{cite}
\usepackage{amsmath,amssymb,amsfonts}
\usepackage{algorithmic}
\usepackage{graphicx}
\usepackage[caption=false, font=footnotesize]{subfig}

\usepackage{textcomp}
\usepackage{xcolor}
\usepackage[nolist]{acronym}
\usepackage{setspace}  
\usepackage[titles,subfigure]{tocloft}  
\usepackage{rotating}  
\usepackage{svg}
\usepackage[normalem]{ulem}  
\usepackage{float}
\usepackage[nolist]{acronym}
\usepackage{booktabs}  
\usepackage{numprint}  
\usepackage{wrapfig}
\usepackage{multicol}
\usepackage{multirow}
\usepackage{lipsum}
\usepackage{import}

\newcommand{\vc}{{\bf c}}
\newcommand{\vx}{{\bf x}}

\newcommand{\vh}{{\bf h}}

\iffalse
\usepackage[backref=section]{hyperref}
\else
\usepackage[backref=page]{hyperref}
\fi

\hypersetup{
    colorlinks=true,
    linkcolor=blue,
    filecolor=blue,      
    urlcolor=blue,
    citecolor=blue,
}

\usepackage[capitalise]{cleveref}

\DeclareGraphicsExtensions{.svg}

\def\BibTeX{{\rm B\kern-.05em{\sc i\kern-.025em b}\kern-.08em
    T\kern-.1667em\lower.7ex\hbox{E}\kern-.125emX}}

\begin{acronym}
\acro{GP}{Gaussian Process}
\acroplural{GP}[GPs]{Gaussian Processes}
\acro{MPPI}{Model-Predictive Path Integral Control}
\acro{MPC}{Model-Predictive Control}
\acro{NN}{Neural Network}
\acro{RNN}{Recurrent Neural Network}
\acro{FNN}{feed-forward Neural Network}
\acro{SGP}{Sparse Gaussian Process}
\acroplural{SGP}[SGPs]{Sparse Gaussian Processes}
\acro{LSTM}{Long Short-Term Memory}
\acro{ADAM}{Long Short-Term Memory}
\end{acronym}

\begin{document}

\title{A Multi-step Dynamics Modeling Framework For Autonomous Driving In Multiple Environments\\
}


\author{Jason Gibson$^{1}$, Bogdan Vlahov$^{1}$, David Fan$^{2}$, Patrick Spieler$^{2}$, Daniel Pastor$^{2}$,\\ Ali-akbar Agha-mohammadi$^{2}$, Evangelos A. Theodorou$^{1}$%
\thanks{$^{1}$Autonomous Control and Decision Systems Lab, Georgia Institute of Technology, Atlanta GA 30313 USA (\textit{Corresponding author: jgibson37@gatech.edu)}} 
\thanks{$^{2} $ NASA Jet Propulsion Laboratory, California Institute of Technology, Pasadena, CA, USA}%
\thanks{This paper has a supplementary video available to view at {\tt\footnotesize https://youtu.be/bzyn-qpRJIk}}
}

\maketitle

\begin{abstract}
Modeling dynamics is often the first step to making a vehicle autonomous.  While on-road autonomous vehicles have been extensively studied, off-road vehicles pose many challenging modeling problems.  An off-road vehicle encounters highly complex and difficult-to-model terrain/vehicle interactions, as well as having complex vehicle dynamics of its own.  These complexities can create challenges for effective high-speed control and planning. In this paper, we introduce a framework for multistep dynamics prediction that explicitly handles the accumulation of modeling error and remains scalable for sampling-based controllers.
Our method uses a specially-initialized \ac{LSTM} over a limited time horizon as the learned component in a hybrid model to predict the dynamics of a 4-person seating all-terrain vehicle (Polaris S4 1000 RZR) in two distinct environments.
By only having the \ac{LSTM} predict over a fixed time horizon, we negate the need for long term stability that is often a challenge when training recurrent neural networks.
Our framework is flexible as it only requires odometry information for labels.
Through extensive experimentation, we show that our method is able to predict millions of possible trajectories in real-time, with a time horizon of five seconds in challenging off road driving scenarios.
\end{abstract}

\acresetall
\section{Introduction}


\ac{MPC} methods have been used for real-time controls applications for many years \cite{pagot2020realtime,kabzan2019learningbased, williams2018information, draeger1995model} in real-time control of complex systems.
These methods are able to handle noisy inputs and uncertainty through fast replanning. 
However, one of the main limiting factors of these approaches is fast, accurate dynamics models.

The most common dynamics model historically has been linear approximations obtained through system identification \cite{survey-learning}.
The inherent nonlinear nature of the world pushed the models towards nonlinear or linearized approximations focused around the core principal of physics-based approximations. 
Unfortunately, these equations are prohibitively difficult to compute, requiring simplifying assumptions \cite{pagot2020realtime}, or rely on parameters that are costly to determine with any certainty \cite{hjallmarsson}.
Therefore, we often have first principal models where nonlinear effects are ignored or greatly simplified \cite{Freddi2011}. 
Even with this limitation, \ac{MPC} methods are able to accomplish challenging real world tasks.
With fast replanning, even significant model error can be mitigated.

Fast replanning induces computational restrictions that are magnified when considering sampling-based controllers, where the number of forward passes on the prediction can exceed millions of computation a second. 
Still, a desire for increased accuracy of dynamics without sacrificing computational efficiency has lead to a prevalence of black-box learning methods in dynamics modeling.
The direction of learned components can be broadly categorized by the following methods: \acp{GP} \cite{Hewing2017,kabzan2019learningbased}, Neural Networks \cite{williams2020locally, draeger1995model}, and others outlined in this survey paper on dynamics model learning \cite{survey-learning}.
Each approach has potential drawbacks for hardware systems in the real world. 
The balance of computation and predictive accuracy is the main limitation when it comes to selecting models for \ac{MPC}. 
Using physics-based equations to integrate predicted accelerations or velocities can simplify the learning problem and improve generalization in various systems \cite{Yu2019} \cite{Singh2019}.
Models using this structure are often called hybrid models.




While there are various architectures and frameworks for the dynamics model in \ac{MPC}, they tend to be used in a very similar fashion, and suffer from two main drawbacks.  
First, trajectories are often computed recursively, using the previous steps' output as the input to the model in the next step.
This inherent recursive nature can lead to significantly decreased accuracy as the time horizon increases, and a small overestimate can quickly blow up a state value and lead to nonsensical trajectories.  
Second, a more nuanced drawback of single-step learning is that lower frequency phenomena are effectively treated as random noise in the learning problem. 
Indeed, prior work has shown the sensitivity of selecting $\Delta t$ for different systems \cite{Liu2020} in single-step prediction. 
Therefore, a good solution to both of these problems is to consider the trajectory prediction problem as a multi-step prediction problem rather than a single-step one.


The recursive nature of how these models are used lends itself toward a model structure that is designed for recurrence. 
In this work, we will look primarily at the \ac{LSTM} architecture \cite{Sak, Hochreiter1997} due to its ability to avoid the vanishing gradient problem often seen in other recurrent structures.
We will follow the example of \cite{Mohajerin2019} and use another network to initialize the hidden and cell state of the \ac{LSTM}.
Our \ac{LSTM} prediction models will only need to be run for a fixed time horizon, so training them for continual stability is not required.

In this work, we demonstrate a hybridization architecture that is successfully able to learn the dynamics of the same vehicle in two diverse environments.
We combine initialized LSTMs \cite{Mohajerin2019} with the integrated loss function \cite{Seegmiller} while keeping the networks small enough to run in real-time on hardware systems.
Through the use of the integrated loss function, we are able to construct a modeling architecture that is robust to changing environments, while only requiring relatively smooth odometry estimates for training. 
This allows seamless integration of human driving data and autonomously controlled data into the training process.
We emphasize that our specific hybrid setup should be viewed as a hyperparameter with extensive room for exploration.
The key is the integrated loss function allowing for predictions with arbitrary transformations and integration.
To our knowledge, this is the first work to combine initialized \acp{LSTM} with the integrated loss function.
The results are validated on a real vehicle using $650,000$ trajectories of driving data.

\section{Related Works}
In the world of autonomous racing, accurate dynamics modeling is crucial to achieving performance at the limits of handling. \cite{pagot2020realtime} uses a kino-dynamic model that improves upon the nominal bicycle model by incorporating drag, road slop effects, understeer/oversteer characteristics, and pose relative to an optimal race line. They achieve real-time performance with this model in an iterative \ac{MPC} solver. However, this model assumes there is no tire slip, which is not applicable in off-road conditions. \cite{berntorp2020trajectory} demonstrates \iac{MPC} approach using a model that learns tire forces as deviations from a library of pre-selected tire models. This allows for control over snow-covered terrain but requires a large body of domain-specific knowledge to generate the library of tire models. Our approach allows for a relaxation on the amount of knowledge required to build the model.

Using learned models for \ac{MPC} to achieve an acceptable balance between accuracy and computational complexity has been investigated for many years.
In 1995, \cite{draeger1995model} used \iac{FNN} to learn the dynamics of the pH in a plant with pump flow rates as the control. They successfully showed that a model trained using a single-step prediction loss and combined with \ac{MPC} could outperform a PI controller  for achieving and maintaining desired pH setpoints. Since then, \cite{williams2016aggressive} has used neural networks trained on single-step prediction loss functions in sampling-based \ac{MPC} controllers to capture nonlinear behavior, such as vehicle drift while turning, leading to improved performance. 

One approach to prevent this recursive error magnification while learning a model is to use \acp{GP}\cite{Hewing2017,kabzan2019learningbased}, whose errors remain bounded when used for predictions which lie near previously collected data points. \cite{kabzan2019learningbased} uses  \acp{SGP} to learn the difference between a nominal bicycle dynamics model and real data gathered from an electric race car. This combination allows them to run an iterative \ac{MPC} solver in real-time and update the model in real-time to new environment information. However, this sort of model is restricted to problems which allow for traversing the same location repeatedly, such as on a race track. \cite{Singh2019} \cite{Qi2019} uses \ac{LSTM} models for prediction of dynamical system inside hybrid models, but none applies the integrated loss function which prevents our hybrid model architecture from being applied.




\section{Model Learning Framework}

\begin{figure*}[ht]
\centering
\includegraphics[width=0.75\textwidth]{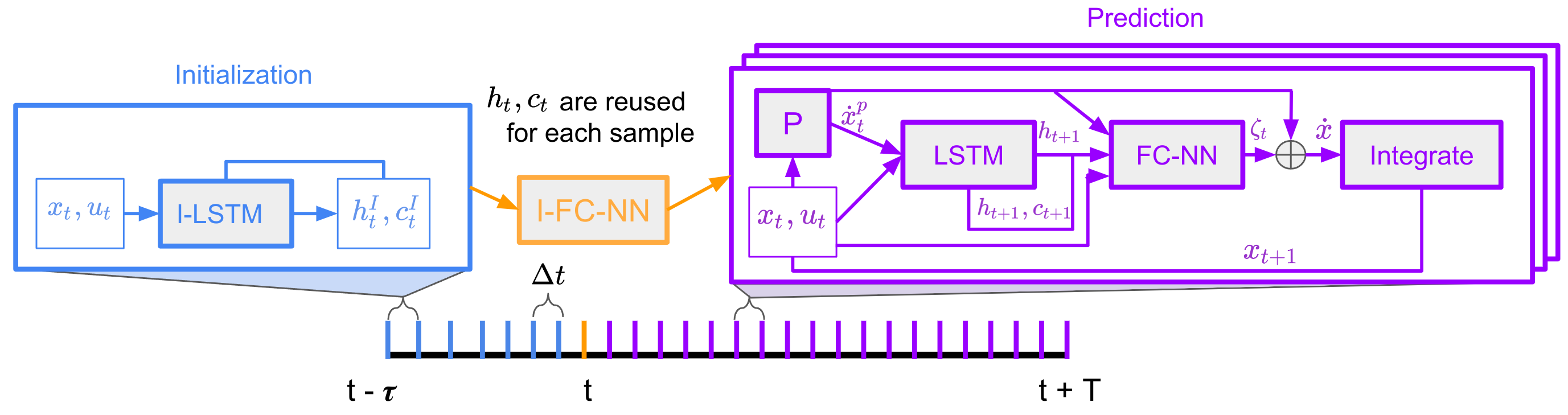}
\caption{Figure showing overall architecture. Note the initialization network is run a single time and $h_t, c_t$ are reused for each trajectory sampled using the prediction step.}
\label{fig:architecture}
\end{figure*}

We will be using the \ac{MPPI} algorithm as a baseline \ac{MPC} approach for computational comparisons.
Ultimately, we want models that can compute 18,432 samples at around 15 Hz. 
The target time horizon of the prediction is $T = 5.0$ with a $\Delta t = 0.02$.
This results in $\approx 69, 000, 000$ forward passes on the dynamics a second.
Our specific implementation of \ac{MPPI} will be able to leverage GPUs through CUDA, as with prior publications \cite{Gandhi2021}.
We specifically target a Nvidia RTX 3080, but these networks can run in real-time on smaller hardware systems as well.
We will follow prior work on hybrid dynamics models \cite{survey-learning}, and predict the derivative of the state we care about and integrate using the relevant kinematic equations.
Our overall architecture can be summarized in \cref{fig:architecture}. 

\textbf{Initialized LSTM}:
We now briefly review \cite{Mohajerin2018} where a new method of initializing \acp{LSTM} was proposed. 
\ac{LSTM} training often has to deal with how to initialize the hidden and cell states of the \ac{LSTM}.
The most common approach with recurrent networks is to define a washout period, where the network is run on prior data for a predefined time period \cite{jaeger2002tutorial}.
Prior work has treated the washout period as a hyperparameter, and effectively loses the data used during the washout period since no loss can be computed on it.
Furthermore, washout can also lead to instability during training \cite{zimmermann_forecasting_2012}.
\cite{Mohajerin2018} treats the problem of initializing the hidden/cell states as part of the training process. 
This is done by having a network predict the initial hidden/cell state of another network using some predefined buffer over history denoted by $\tau$. 

The followup paper \cite{Mohajerin2019} primarily focuses on exploring what type of initialization network is best. 
It does a robust comparison of washout, \iac{FNN}, and \iac{RNN}.
Their extensive results with very large networks indicate that \ac{RNN}s are the best suited architecture.
They also investigate and demonstrate that hybrid architectures are better able to predict farther into the future with fewer weights.
We will continue this analysis for the autonomous driving setting in off-road environments.
The general initialized network architecture can be shown in the following equations.
\begin{align}
    \begin{bmatrix} \vc_{t-\tau}^I & \vh_{t-\tau}^I \end{bmatrix}^T &= \kappa_I(\vx_{t-\tau}, \mathbf{0}, \mathbf{0}) \label{eq:init-lstm-1}\\
    \vdots \nonumber\\
    \begin{bmatrix} \vc_t^I & \vh_t^I \end{bmatrix}^T &= \kappa_I(\vx_t, \vc_{t-1}^I, \vh_{t-1}^I)\label{eq:init-lstm-2}\\
    \begin{bmatrix} \vc_t^P & \vh_t^P \end{bmatrix}^T &= \vartheta_I(\vh_{t}^I, \vx_t)\label{eq:init-lstm-3}\\
    \begin{bmatrix} \vc_{t+1}^P & \vh_{t+1}^P \end{bmatrix}^T &= \kappa_P(\vx_t, \vc_t^P, \vh_t^P)\label{eq:pred-lstm-1}\\
    \zeta_{t} &= \vartheta_P(\vh_{t+1}^P, \vx_t) \label{eq:pred-lstm-2}
\end{align}
Where $\kappa_I$ represents the \ac{LSTM} initializer network, $\vartheta_I$ is the \ac{FNN} initializer output network, $\kappa_P$ represents the \ac{LSTM} prediction network, $\vartheta_P$ represents the \ac{FNN} prediction output network, and $\vh_t^I, \vc_t^I, \vh_t^P, \vc_t^P, \vx_t, \zeta_t,$ are the initializer hidden and cell vectors, prediction hidden and cell vectors, state vector, and output vector respectively at time $t$.
A key note is that the initializer network only runs on historical data and has no information about the future controls of the vehicle.
This means sampling-based approaches can reuse the same initialization output for all samples.
Furthermore, there is no restriction on the input to the initializer networks; it can use a different input vector than the predictor network.

\textbf{Integrated Loss}:
The integrated loss function penalizes the deviation from the true state trajectory in position mainly \cite{Seegmiller}. 
In essence this penalizes the accuracy of the integration as well as the networks prediction. 
The main benefits of \cite{Seegmiller} can be summarized by the following points.
First, we only require smooth position estimates, so there is no need to differentiate noisy odometry readings to get accelerations or approximate outputs that are difficult to compute.
The input to the loss function is only position estimates so learned models can predict arbitrary values which would be challenging to determine labels for.
We use this in our hybrid architecture to predict drag forces on wheels in the wheel frame.
Second, we explicitly penalize accumulated error which allows for stable long term predictions far into the future that can account for slowly evolving changes to the dynamics.
The two main drawbacks of this method is that we can get non-smooth outputs with low loss in the first couple predictions and increased loss function complexity during learning.
The first is an ongoing problem in this work, and the second is mitigated by using modern libraries for backpropagation \cite{PyTorch}.

\section{Hybrid Model Setup}
\label{sec:hybrid}

The vehicle dynamics that we learn are for the Polaris S4 1000 RZR.
It is fit with a drive-by-wire kit that enables autonomous driving as well as logging of control inputs from a human driver.
The control inputs for the vehicle are desired throttle, brake, and steering wheel position.
During autonomous operation, a physical actuator must turn the steering wheel and actuate the brake, which introduces significant delay.
The delay on throttle actuation is very small and can safely be ignored.
These delays are not present in the human driving data, and the human is able to actuate these controls far faster than the autonomous system.
This poses a problem as human driving data is cheaper to attain, and the vehicle can be driven more aggressively.
In order to make use of this data, we create a model architecture that allows for the decoupling of the steering wheel angle and the brake commanded from the prediction of the body accelerations.
The inputs to our models consist of position and velocity coming from odometry and an approximation of roll and pitch derived from an elevation map.
Our model is broken into 3 main components summarized below.

The first two components predict the brake and steering actuation delay.
These are trained only on autonomous data and use ground-truth odometry and control commands as inputs.
Both systems can be modeled parametrically by a rate-limited first-order lag in addition to a neural network as outlined below,
\begin{align}
    p_b &= \min\left(\max\left(\left(u^b_t - b^u_t\right) C_B, -\dot{b}^u_{lim} \right), \dot{b}^u_{lim} \right) \\
    \zeta^b_t& = \vartheta_P^b\left(b^u_t, u^b_t, p_b\right)\\
    b^u_{t+1} &= b^u_t + \left(p_b + \zeta^{b}_t\right) \Delta t
\end{align}
Where $p_b$ is the parametric approximation of the model with rate limit $\dot{b}^u_{lim}$ and constant $C_B$, $u^b_t$ is the commanded brake, and $b^u_t$ is the current brake state.
$\zeta^b_t$ is the output of the network, the variables after that denote the inputs. The steering delay is modeled with the following equations,
\begin{align}
    p_\delta &= \min\left(\max\left(\left(u^\delta_t - \delta_t\right) C_\delta, -\dot{\delta}^u_{lim} \right), \dot{\delta}^u_{lim} \right) \\
    \zeta^\delta_t &= \vartheta_P^\delta\left(v^x_t, u^\delta_t, \delta_t, \dot{\delta}_t, p_\delta\right) \\
    \delta_{t+1} &= \delta_t  + \left(p_\delta + \zeta^\delta_t\right) \Delta t
\end{align}
where the notation matches the above but with $\delta$ as the steering angle position and $v^x_t$ denotes the body frame forward velocity of the vehicle. 
The parameters used in the above equations are fit using the Adam optimizer \cite{Adam2014} on the purely parametric version.
The loss function is mean squared error on the state ($\delta, b^u_t$).

\begin{figure}[ht]
\centering
\includegraphics[width=0.85\columnwidth]{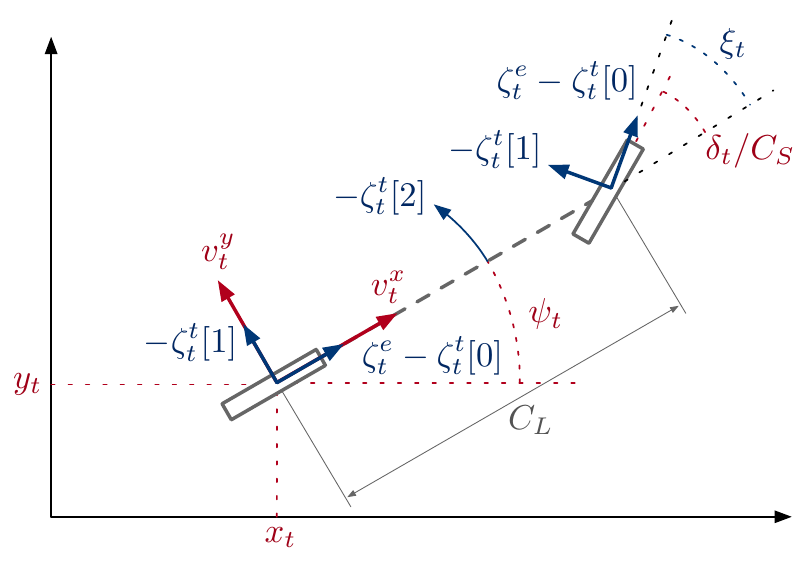}
\caption{Hybrid model coordinate frame definitions}
\label{fig:wheelFrames}
\end{figure}

The third component is made up of two different networks.
An engine model predicts the force imparted by the engine on the wheels, $\zeta^e_t$, and a terradynamics model predicts an environment-dependent drag force, $\zeta^t_t$.
The coordinate frames are pictured in \cref{fig:wheelFrames}.
We normalize by the mass of the vehicle as it is constant between every equation.
The outputs of the networks are scaled by 10, which was chosen to be the maximum expected acceleration value output from the network.
This reduces the expected output range to be $[-1, 1]$ which prevents issues when using $\tanh$ activation functions in the output network.

The steering, brake, and engine model are all shared across each environment, but the terradynamics model is switched out dependent on what terrain we are on.
Initially, we train a single terradynamics network, then allow specialization farther into the learning process.
We assume the environment class is known \emph{a priori}, and the correct model is loaded at run time. 
The environment is a broad definition like desert, wooded areas, etc.
We feel this is reasonable to predict from perceptual inputs in future work, and is not restrictive. 
The combined equations are outlined below with $g = 9.81 m/s^2$,
\begin{align}
    \xi_t &= \tan\left(\delta_t / C_W\right)\\
    p^\psi_t &= \frac{v^x_t}{C_L} \tan\left(\frac{\delta_t}{C_S}\right) \label{eq:steering_rate_parametric}\\
    \zeta^e_t &= \vartheta_P^e\left(v^x_t, u^t_t, b^u_t\right)\\
    \zeta^t_t &= \vartheta_P^t(v^x_t, v^y_t, \dot{\psi}_t, \delta_t, \dot{\delta}_t, g\sin\left(\phi_t\right), g\sin\left(\theta_t\right), p^\psi_t, \xi_t, \zeta^e_t)\\
    \dot{v}^x_t &= \zeta^e_t - \zeta^t_t[0] + \left(\zeta^e_t - \zeta^t_t[0]\right)\cos\left(\xi_t\right) + \zeta^t_t[1] \sin\left(\xi_t\right) \\
    \dot{v}^y_t &= - \zeta^t_t[1] - \zeta^t_t[1] \cos\left(\xi_t\right) + \left(\zeta^e_t - \zeta^t_t[0]\right) \sin\left(\xi_t\right)\\
    \dot{\psi}^t_t &= p^\psi_t - \zeta^t_t[2] 
\end{align}
where $u^t_t$ is the commanded throttle, $\phi_t$ is the vehicle roll angle, $\theta_t$ the pitch angle and $v^x_t$ and $v^y_t$ are the body longitudinal and lateral velocity.
We integrate the above network predictions as detailed below,
\begin{align}
    \label{eq:kinematics-1}
    x_{t+1} &= x_t + \left(v^x_t \cos(\psi) - v^y_t \sin(\psi)\right) \Delta t \\
    \label{eq:kinematics-2}
    y_{t+1} &= x_t + \left(v^x_t \sin(\psi) + v^y_t \cos(\psi)\right) \Delta t \\
    \label{eq:kinematics-3}
    \psi_{t+1} &= \psi_t + \dot{\psi}^t_t \Delta t \\
    \label{eq:kinematics-4}
    v^x_{t+1} &= v^x_t + \dot{v}^x_t \Delta t \\
    \label{eq:kinematics-5}
    v^y_{t+1} &= v^y_t + \dot{v}^{y}_t \Delta t 
\end{align}
The parametric constants are fit first using Adam on the full dataset to get the best pure parametric approximation, but are allowed to change while training the network.
Typically, these models remain within $5\%$ of the initial value during training.
$C_S$ and $C_W$ are very similar constants that differ in that $C_S$ was fit to minimize error on $\psi$ using ground truth values and \cref{eq:steering_rate_parametric}. 
$C_W$ was measured to correctly approximate the angle of the wheel under various steering conditions.

\section{Experimental Setup}

The vehicle used to generate all data is a modified Polaris S4 1000 RZR, as described in \cref{sec:hybrid}.  The vehicle is modified with onboard sensors, compute, and power, and is designed to be fully autonomous and operational over multi-kilometer off-road traversals.  The onboard compute consists of an AMD Threadriper 3990X, 8x 32GB DDR4 RAM, and 4x NVidia RTX 3080's with 10GB of RAM each (only one RTX 3080 is allocated for MPC compute).  Onboard sensors include 3x LiDARs (2x front-facing, 1x rear-facing, Velodyne VLP-32C), 4x Stereo/RGB cameras (3x front, 1x rear, CRL S27), and other various sensors which are not used in this work.  The vehicle is also equipped with IMUs, GPS, and wheel and suspension encoders. 

\textbf{Inputs}: We briefly describe the two external inputs to the dynamics model, namely, odometry and elevation map.  Odometry is generated using the approach described in \cite{Fakoorian2022}.  Inputs to odometry include IMU and LiDAR data.  A factor-graph based approach is used to fuse various sources of measurements, including IMU integration and LIO-SAM \cite{LIOSAM}, to generate a low latency, high accuracy odometry update at 400Hz with $<$50ms of delay, suitable for capturing the aggressive motions of the vehicle.

The elevation map is generated using a geometric and semantic-based approach which utilizes LiDAR and camera data to create a dense, locally consistent map of the environment.  In this work, we require a reliable ground height and slope estimate from sensor data.  This problem can be challenging, especially in environments with various degrees of vegetation, which is both compressible and obscuring.  Relying on a purely geometric map from aggregated LiDAR scans is insufficient for predicting where the tires of the vehicle will settle as the vehicle traverses the terrain.  Building on prior work (see \cite{Fan2021}), we take a combined semantic and geometric approach where semantic cues are used to remove compressible vegetation from ground height estimates.  Here we briefly describe the approach, though further detail is beyond the scope of this work.  First, odometry is used to align LiDAR pointclouds and camera images.  Semantic classification is performed on this data to identify grass, bushes, trees, and other compressible elements of the environment.  LiDAR is aggregated into a local map using a GPU-based raytracing and voxel mapping approach \cite{GVOM}.  An initial guess of elevation is generated using the minimum heights of the voxel map.  This guess is refined using a combination of variance-weighted averaging, semantic-based heuristics, and interpolation.

\begin{figure}[ht]
    \subfloat[Helendale\label{fig:helendale}]{\includegraphics[width=0.49\linewidth]{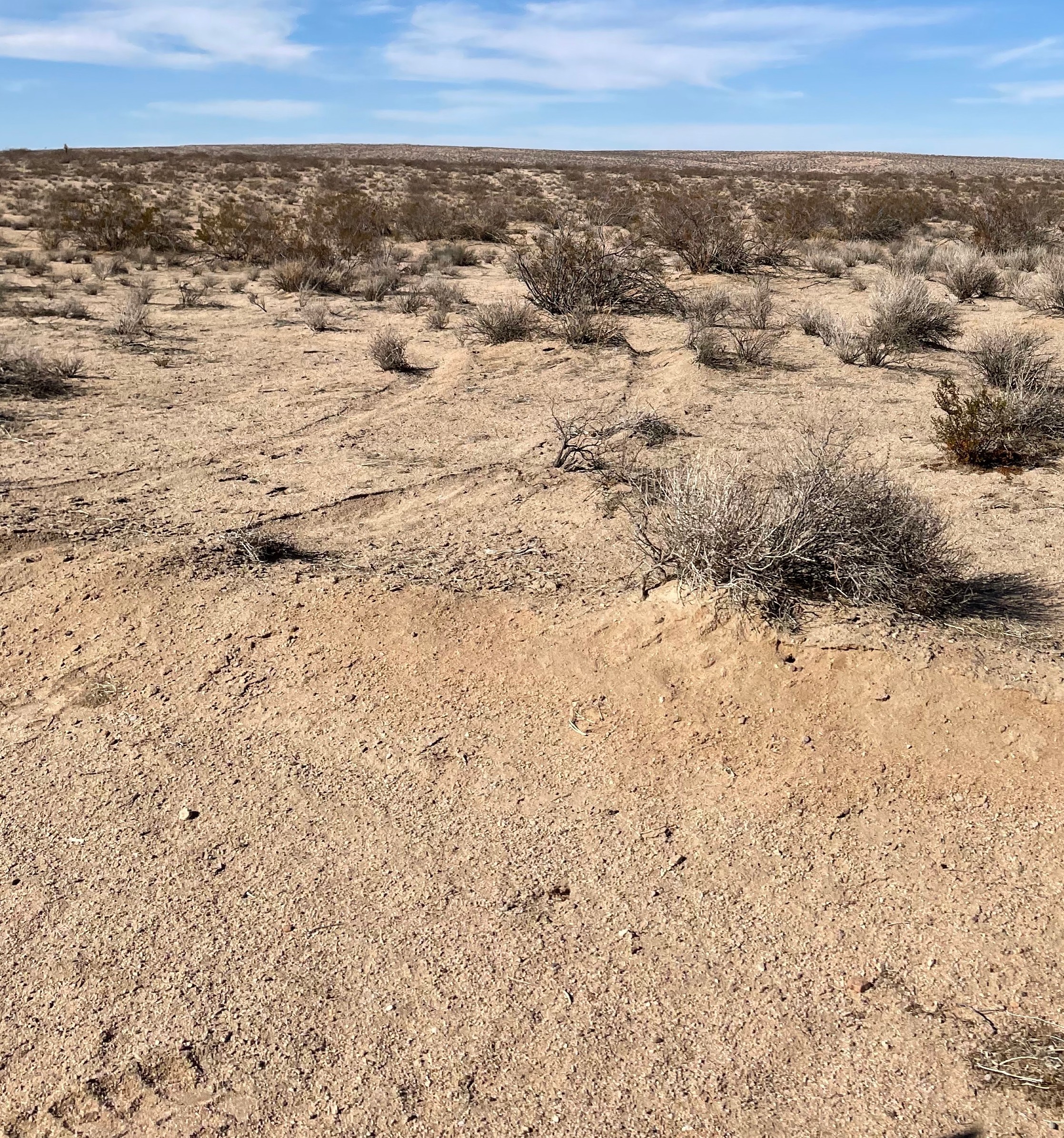}}
  \hfill 
  \subfloat[Paso Robles\label{fig:halter}]{\includegraphics[width=0.49\linewidth]{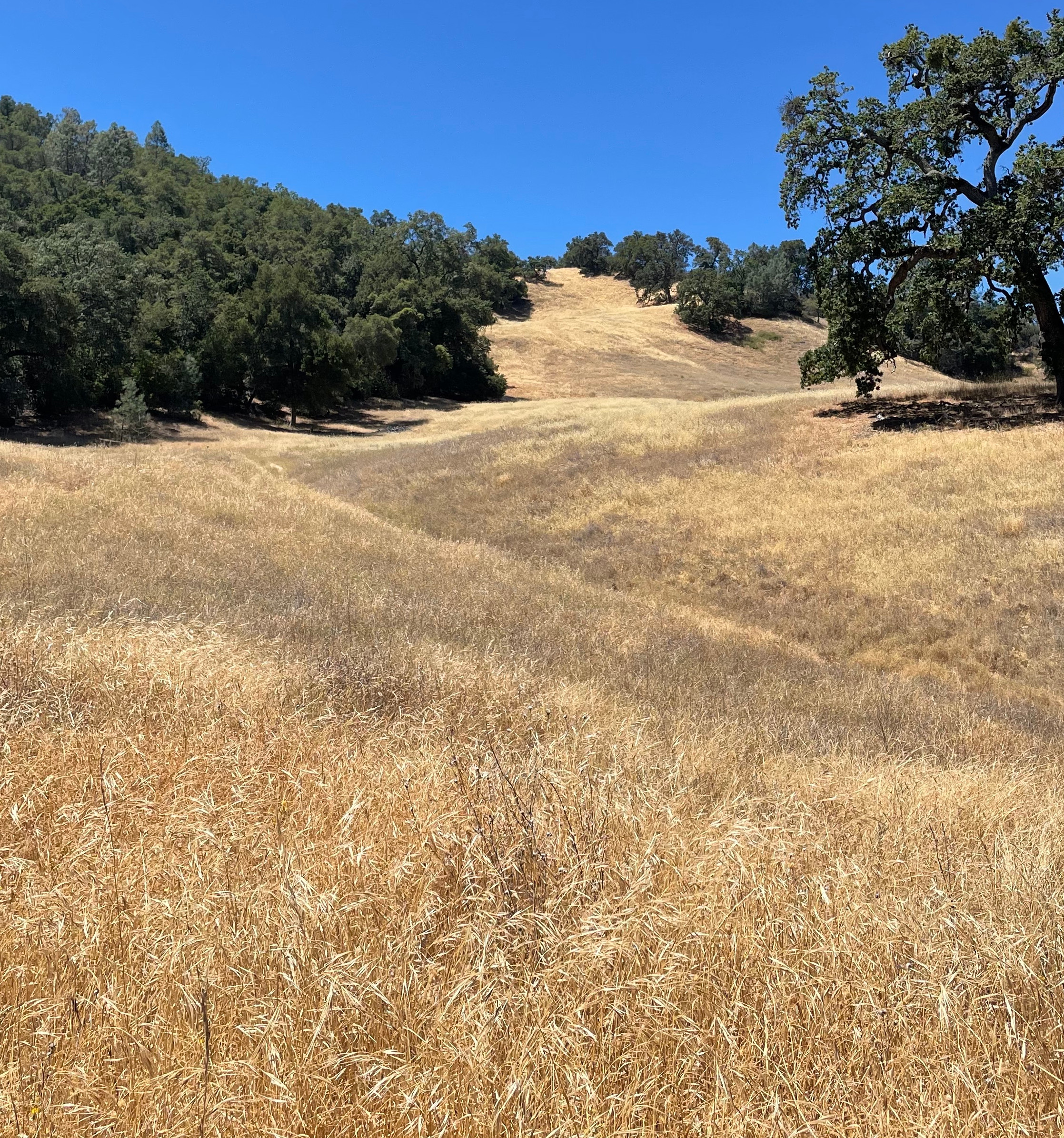}}
\caption{Pictures from the two environments where the datasets were collected.}
\end{figure}

\textbf{Helendale}: The first dataset consists of on-trail and off-road driving near Helendale CA, located in the Mojave Desert (Fig. \ref{fig:helendale}).
The specific area of the dataset includes numerous hills around 10 meters tall, and is very dry with loose sand.
The dataset includes $\approx 300,000$ trajectories of human driving data and $\approx 100,000$ trajectories of autonomous driving data.

\textbf{Paso Robles}: The second dataset is from trails and off-road driving near Paso Robles, CA.
It is a primarily consists of dry grassland and oak woodland, denoted as the chaparral environment (Fig. \ref{fig:halter}).
This subsection of data has large, steep slopes of up to $40^\circ$ inclination and more than 100~m tall.
The ground is entirely covered with dense, dry grass where the vehicle easily loses traction.
Our dataset includes $\approx 100,000$ trajectories of human driving and $\approx 150,000$ trajectories of autonomous driving data.

\begin{table*}[hbtp!]
\footnotesize
\centering
\caption{Mean error and first standard deviation of error for combined and combined autonomous}
\begin{tabular}{cccccccc}
\toprule
\textbf{\begin{tabular}[c]{@{}c@{}}Data \\ Type\end{tabular}} &
  \textbf{Network} &
  \textbf{\begin{tabular}[c]{@{}c@{}}Distance \\ Error (m)\\ t = 1\end{tabular}} &
  \textbf{\begin{tabular}[c]{@{}c@{}}Yaw \\ Error (rad)\\ t = 1\end{tabular}} &
  \textbf{\begin{tabular}[c]{@{}c@{}}Distance \\ Error (m)\\ t = 5\end{tabular}} &
  \textbf{\begin{tabular}[c]{@{}c@{}}Yaw \\ Error (rad)\\ t = 5\end{tabular}} &
  \textbf{\begin{tabular}[c]{@{}c@{}}Distance \\ Error (m)\\ t = 10\end{tabular}} &
  \textbf{\begin{tabular}[c]{@{}c@{}}Yaw \\ Error (rad)\\ t = 10\end{tabular}} \\
\midrule
\multirow{5}{*}{\begin{tabular}[c]{@{}c@{}}C\end{tabular}} &
  \multicolumn{1}{c|}{P} &
  $0.377 \pm 0.127$ &
  \multicolumn{1}{c|}{$0.036 \pm 0.002$} &
  $4.203 \pm 10.461$ &
  \multicolumn{1}{c|}{$0.175 \pm 0.054$} &
  $11.153 \pm 77.551$ &
  $0.339 \pm 0.184$ \\
 &
  \multicolumn{1}{c|}{K-LSTM} &
  ${\bf 0.270} \pm 0.088$ &
  \multicolumn{1}{c|}{${\bf 0.033 \pm 0.001}$} &
  $1.845 \pm 2.309$ &
  \multicolumn{1}{c|}{$0.117 \pm {\bf 0.013}$} &
  $5.023 \pm 18.833$ &
  $0.208 \pm 0.046$ \\
 &
  \multicolumn{1}{c|}{KI-LSTM} &
  $0.277 \pm 0.087$ &
  \multicolumn{1}{c|}{$0.036 \pm 0.001$} &
  $1.975 \pm 2.476$ &
  \multicolumn{1}{c|}{${\bf 0.113} \pm 0.014$} &
  ${\bf 4.931} \pm 18.095$ &
  ${\bf 0.195 \pm 0.045}$ \\
 &
  \multicolumn{1}{c|}{H-LSTM} &
  $0.284 \pm 0.096$ &
  \multicolumn{1}{c|}{$0.042 \pm 0.001$} &
  $1.995 \pm 2.708$ &
  \multicolumn{1}{c|}{$0.138 \pm 0.021$} &
  $5.459 \pm 21.833$ &
  $0.249 \pm 0.074$ \\
 &
  \multicolumn{1}{c|}{HI-LSTM} &
  $0.278 \pm {\bf 0.078}$ &
  \multicolumn{1}{c|}{$0.049 \pm 0.002$} &
  ${\bf 1.765 \pm 2.021}$ &
  \multicolumn{1}{c|}{$0.132 \pm 0.021$} &
  $4.981 \pm {\bf 17.143}$ &
  $0.229 \pm 0.065$ \\ \hline
\multicolumn{1}{c}{\multirow{5}{*}{\begin{tabular}[c]{@{}c@{}} CA \end{tabular}}} &
  \multicolumn{1}{c|}{K-LSTM} &
  $0.539 \pm 0.136$ &
  \multicolumn{1}{c|}{$0.239 \pm 0.026$} &
  $7.259 \pm 28.837$ &
  \multicolumn{1}{c|}{$0.775 \pm 0.307$} &
  $18.159 \pm 202.346$ &
  $1.055 \pm 0.520$ \\
\multicolumn{1}{l}{} &
  \multicolumn{1}{c|}{KI-LSTM} &
  $0.472 \pm 0.109$ &
  \multicolumn{1}{c|}{$0.177 \pm 0.015$} &
  $7.040 \pm 21.287$ &
  \multicolumn{1}{c|}{$0.813 \pm 0.0307$} &
  $21.015 \pm 155.293$ &
  $1.431 \pm 0.688$ \\
\multicolumn{1}{l}{} &
  \multicolumn{1}{c|}{H-LSTM} &
  $0.376 \pm 0.071$ &
  \multicolumn{1}{c|}{${\bf 0.036} \pm {\bf 0.001 }$} &
  $2.465 \pm 3.737$ &
  \multicolumn{1}{c|}{$0.149 \pm 0.016$} &
  $6.239 \pm 26.865$ &
  $0.276 \pm {\bf 0.049 }$ \\
\multicolumn{1}{l}{} &
  \multicolumn{1}{c|}{HI-LSTM} &
  ${\bf 0.272} \pm {\bf 0.042}$ &
  \multicolumn{1}{c|}{$0.040 \pm 0.001$} &
  ${ \bf 2.089 } \pm {\bf 3.463 }$ &
  \multicolumn{1}{c|}{${ \bf 0.145 } \pm { \bf 0.016 }$} &
  ${ \bf 5.722 } \pm {\bf 22.936 }$ &
  ${ \bf 0.268 } \pm 0.051$
\end{tabular}
\label{tab:full_results}
\vspace{-4mm}
\end{table*}
\raggedbottom


\section{Results}

Throughout this results section, we will be covering results on the RZR vehicle trained with a time horizon of $T=5$s and a $\Delta t = 0.02$s.
The main results are summarized in \cref{tab:full_results}.
This compares 5 different architectures of models on the total combined dataset.
The overall training set was approximately $\approx 500,000$ trajectories, and a test set of $\approx 150,000$ trajectories.
There is potential overlap within the same trajectory since a trajectory going from $[t_0, t_T]$ could include values from another trajectory from $[t_1, t_{T+1}]$.
Overlap is constrained within the validation or test set.
These two sets are pulled from the same day of testing, but are otherwise independent.

\emph{H} represents the hybrid model outlined in \cref{sec:hybrid}, whereas \emph{K} represents the kinematic only model, and \emph{I} denotes use of an initializer network for prediction, such that
\emph{HI-LSTM} is an hybrid architecture with an initialization network.
If there is no initializer network the initial hidden/cell state of the predictor network is set to zero.
The $K$ networks directly predict the outputs $\dot{v}^x_t, \dot{v}^y_t, \dot{\psi}$ without the hybrid architecture.
\emph{K-LSTM} is a \ac{LSTM} with no initializer network and only uses input from odometry and elevation map, but still uses the kinematic \cref{eq:kinematics-1,eq:kinematics-2,eq:kinematics-3,eq:kinematics-4,eq:kinematics-5} for integration.
This restricts \emph{K-LSTM} to using $(v^x_t, v^y_t, \dot{\psi}_t, \delta_t, \dot{\delta}_t, \phi_t, \theta_t, u^t_t, u^b_t, \eta_t)$ as inputs, where $\eta_t$ is -1 for Paso Robles and 1 for Helendale.
This means that the \emph{K}-networks do have access to what environment they are being run on.
The initialization and prediction networks take identical input values.

Our networks are implemented in PyTorch \cite{PyTorch} for training and comparisons on data.
The real-time results are in a custom implementation of an \ac{LSTM} network written in C++/CUDA.
All networks are trained for $10$ epochs, with the last $3$ epochs allowing for specialization of the terradynamics network to each environment.
The network size for each architecture is given in the following notation:
\begin{center}
[$\kappa_I$ input dim][$\kappa_I$ hidden dim]$\times$[$\vartheta_I$ architecture] $-$ \\ 
{[$\kappa_P$ input dim][$\kappa_P$ hidden dim]$\times$[$\vartheta_P$ architecture]}.
\end{center}
Where the notation follows that used in \cref{eq:init-lstm-1,eq:init-lstm-2,eq:init-lstm-3,eq:pred-lstm-1,eq:pred-lstm-2}.
For the output networks, $\vartheta$, the numbers represent the size of the layers.
For example $5,10,3$ would mean \iac{FNN} that takes in 5 inputs, has a middle layer with 10 neurons with a $\tanh$ activation, and outputs 3 values.

\begin{itemize}
    \item \emph{K-LSTM} and \emph{KI-LSTM}: 
    \begin{itemize}
        \item Delay: $[2][60]\times[62, 100, 10] - [2][5]\times[7, 10, 1]$
        \item Steering: $[4][60]\times[64, 100, 10] - [4][5]\times[9, 5, 1]$
        \item Engine/Terra Combo: $[10][60]\times[70, 100, 60] - [10][30]\times[40, 40, 3]$
    \end{itemize}
    \item \emph{H-LSTM} and \emph{HI-LSTM}: 
    \begin{itemize}
        \item Delay: $[2][60]\times[62, 100, 10] - [3][5]\times[8, 10, 1]$
        \item Steering: $[4][60]\times[64, 100, 10] - [5][5]\times[10, 5, 1]$
        \item Engine: $[3][60]\times[63, 100, 10] - [3][5]\times[8, 20, 1]$
        \item Terra: $[10][60]\times[70, 100, 20] - [10][10]\times[20, 20, 3]$
    \end{itemize}
\end{itemize}

Notice that the combined engine and terradynamics network in the \emph{K} versions has significantly more weights $\approx 3X$.
The increase in weights was to compensate for the lack of a parametric approximation to predict around.

A key part of optimizing performance in CUDA code is reducing access to memory in the parallel threads.
Some additional memory beyond the weights must be used in order to parallelize the computation of the network.
As the initializer network only needs to be run once for all samples, it can remain on the CPU. 
Only the predictor network needs to run in parallel on the GPU.
The memory usage and runtimes of the models in CUDA are outlined in \cref{tab:computation}.
Notice that the amount of memory required for computing the kinematic networks greatly exceeds that needed by the hybrid architecture while giving similar results. 

The parametric model used as a baseline comparison is a linear approximation of the engine and terradynamics model combined with the parametric components outlined in \cref{eq:kinematics-1,eq:kinematics-2,eq:kinematics-3,eq:kinematics-4,eq:kinematics-5}, using \cref{eq:steering_rate_parametric} for the yaw rate.
This model does not have any $v^y_t$, just a linear approximation of $\dot{v}^x_t$,
\begin{align}
    \dot{v}^x_t &= C_T u^t_t + C_B b^u_t - C_V v^x_t
\end{align}
where $C_T = 4.0, C_B = 10, C_V = 0.2$.
This was the primary dynamics model used during collection of the autonomous driving data, and performs reasonably well with \ac{MPPI}.

\begin{table}[]
\centering
\footnotesize
\caption{Average CUDA runtimes of 1 MPPI iteration \\
with 18432 samples on a Nvidia RTX 3080}
\begin{tabular}{ccc}
\toprule
Architecture & \begin{tabular}[c]{@{}c@{}}CUDA\\ Memory (kB)\end{tabular} & \begin{tabular}[c]{@{}c@{}}CUDA\\ Runtime (ms)\end{tabular} \\ \midrule
P            & 0.024 & 23.327                                                           \\ 
K-LSTM and KI-LSTM         & 62.468 & 83.708                                                           \\
H-LSTM and HI-LSTM       & 38.08 & 59.346                                                           \\
\bottomrule
\end{tabular}
\label{tab:computation}
\vspace{-4mm}
\end{table}

\textbf{Engine/Terradynamics Performance}: For a comparison independent of the learned actuator delays, we compared our models with ground truth values for the brake state $b^u_t$, steering angle $\delta$ and steering rate $\dot{\delta}$.
The models were run on combined data from autonomous driving and human driving.
Results are in \cref{tab:full_results} under Combined.
All models perform relatively similarly with ground truth values, most resulting in positional errors around 5 meters with a 10 second trajectory.
This shows that the \ac{LSTM} networks are able to predict stable trajectories twice their training horizon.
For engine/terradynamics modeling, the hybrid architecture did not seem to improve performance significantly over the learned networks, indicating that the hybrid architecture was likely not assistive with the learning.
The design space of the hybrid architecture and integration is broad, a different setup could result in improved performance. 
Furthermore, the initializer network had a limited effect on accuracy. 
Increasing the size of the initialization network would likely improve accuracy, a direction for future work would be a detailed analysis of how varying the size effects the accuracy.

\textbf{Entire System Performance}: In this section, we reuse the same networks as previously, but include matching architectures for the brake and steering delay models.
Therefore, the \emph{K} networks will have an \ac{LSTM} directly predict $\dot{b}^u_t$ and $\dot{\delta}$ without any parametric assistance.
This focuses only on the autonomy dataset, a subset of $\approx 50,000$ trajectories for a test set.
Here we see a stark contrast between the hybrid approach and the kinematic approach. 
The brake and steering networks perform worse than their parametric counterparts in these cases where a black box function can be hard to predict.
On average, they have about 2X the average error in steering and very similar prediction accuracy for the brake models.
The brake prediction error is deceptive though, since most trajectories have a constant brake state of zero and no brake applied at all.
Looking at the individual state trajectories, we see slight perturbations from the network that can impact performance significantly. 

The worse steering and brake prediction performance leads to a large increase in the yaw error over time, giving very large distance errors at higher speeds.
Notably, the outputs tend to be more jagged, when previously they were smooth.
It is likely that the network was relying on the perfect values for $\delta, \dot{\delta}, b^u_t$.
Once these values started to become inaccurate due to the modeling errors, the network entered sequences of inputs that were not encountered during training and gave poor results.
The hybrid architecture reduces the reliance by providing a reasonable estimate that is independent of the internal state of the \ac{LSTM}.

The network tends to oscillate the predictions from large positive values to large negative values in a single time step on all values.
$\dot{\psi}$ and $\dot{v}_x$ are prone to larger oscillations, likely due to more noise in labels from odometry. 
The methodology is sensitive to drift and the small jumps in odometry that come from the constraint of delayed LIO optimization with IMU preintegration.
Logically, the integrated loss function will not penalize discontinuous looking state trajectories that match well with position from odometry.
drift is treated as potential movement the network needs to predict, therefore we often see short dynamically infeasible trajectories at velocities close to zero.  
This was somewhat reduced in the trajectory through adding additional losses on velocity, but further investigation into the effect of odometry noise is needed.
In general the models show great ability to predict stable outputs into the future but tend to over smooth the outputs.
Since odometry noise in velocities can result in incorrect positions if just directly integrated, we only penalize the loss of velocity when the error is large. 


\section{Conclusion}
In this work, we propose a framework for dynamics model learning in challenging off-road environments.
Our method combines specially initialized \ac{LSTM} networks with an integrated loss function to allows for flexible learned outputs.
Through hundreds of thousands of trajectories we are able to show generalization across two distinct off-road environments. 
Our method is computationally efficient and can run in real time on our hardware in a sampling based \ac{MPC} approach.

\section*{Acknowledgments}
The research was partially carried out at the Jet Propulsion Laboratory, California Institute of Technology, under a contract with the National Aeronautics and Space Administration (80NM0018D0004).

\bibliography{references}  
\bibliographystyle{ieeetr}

\end{document}